\documentclass[letterpaper, 10 pt, conference]{ieeeconf}
\IEEEoverridecommandlockouts
\usepackage{graphicx}
\usepackage{epsfig}
\usepackage{times}
\usepackage{amsmath}
\usepackage{amssymb}
\usepackage{booktabs}
\usepackage{subcaption}
\usepackage{multirow}
\usepackage{algorithm}
\usepackage{algorithmic}
\usepackage{makecell}

\title{\LARGE \bf
Deep Reinforcement Learning based Local Planner for UAV Obstacle Avoidance using Demonstration Data 
}

\author{Lei He$^{1,3}$, Nabil Aouf$^{2}$, James F. Whidborne$^{3}$ and Bifeng Song$^{1}$
\thanks{*This work was supported by China Scholarship Council. (No. 201806290175)}
\thanks{$^{1}$Lei He and Bifeng Song are with the School of Aeronautics, Northwestern Polytechnical University, Xi'an, 710072, China 
        {\tt\small heleidsn@gmail.com, bfsong@nwpu.edu.cn} }%
\thanks{$^{2}$Nabil Aouf is with the Dept of Electrical and Electronic Engineering, City, University of London, London EC1V 0HB, UK 
        {\tt\small nabil.aouf@city.ac.uk}}%
\thanks{$^{3}$Lei He and James F. Whidborne are with the Centre for Aeronautics,  Cranfield University, Bedfordshire, MK43 0AL, UK
        {\tt\small lei.he|j.f.whidborne}@cranfield.ac.uk}%
}

\begin{document}
	
\maketitle
	
\begin{abstract}
In this paper, a deep reinforcement learning (DRL) method is proposed to address the problem of UAV navigation in an unknown environment. However, DRL algorithms are limited by the data efficiency problem as they typically require a huge amount of data before they reach a reasonable performance. To speed up the DRL training process, we developed a novel learning framework which combines imitation learning and reinforcement learning and building upon Twin Delayed DDPG (TD3) algorithm. We newly introduced both policy and Q-value network are learned using the expert demonstration during the imitation phase. To tackle the distribution mismatch problem transfer from imitation to reinforcement learning, both TD-error and decayed imitation loss are used to update the pre-trained network when start interacting with the environment. The performances of the proposed algorithm are demonstrated on the challenging 3D UAV navigation problem using depth cameras and sketched in a variety of simulation environments. 
\end{abstract}

\section{Introduction}
Unmanned Aerial Vehicles (UAVs) have shown great promise in recent years because of its excellent mobility and flexibility. More and more missions involve navigating through the unknown environment, such as search and rescue. Obstacle avoidance is an essential feature for UAVs to navigate autonomously in a complex environment. However, due to the limited capacity and computing resource, autonomous navigation is still a challenging task.

Conventional robotics methods for exploration and navigation, such as Simultaneous Localisation and Mapping (SLAM), tackle the navigation problem through an explicit focus on position inference and mapping \cite{gao2016online}. However, it requires a large amount of computation and memory resource.  There are also some local approaches which do not need to build a map but act on the sensor data gathered at the current time step directly, such as 3DVHF+ \cite{vanneste20143dvfh+}, potential field method (PFM) \cite{mac2016improved} and other reactive methods \cite{ruf2018real,escobar2018r}. These algorithms are faster but usually unable to find the optimal path.

Recently, some end-to-end methods have been proposed to address the UAV navigation problem. The control command is generated from a trained neural network using raw sensor data directly. Compared with the traditional hierarchical pip-line, deep neural network does not need artificial feature extraction and can deal with high dimension raw sensor data such as images. Also, it runs in a reactive manner without any optimization or searches which is beneficial for real-time application. Deep reinforcement learning (DRL) is usually used to train this end-to-end policy network. However, DRL is sample inefficient which relies on a large amount of interaction data with the environment. Learning from scratch is time consuming and severally limits the application of DRL to many real-world tasks.

\begin{figure}[t]
        \centering
        \includegraphics[width=8cm]{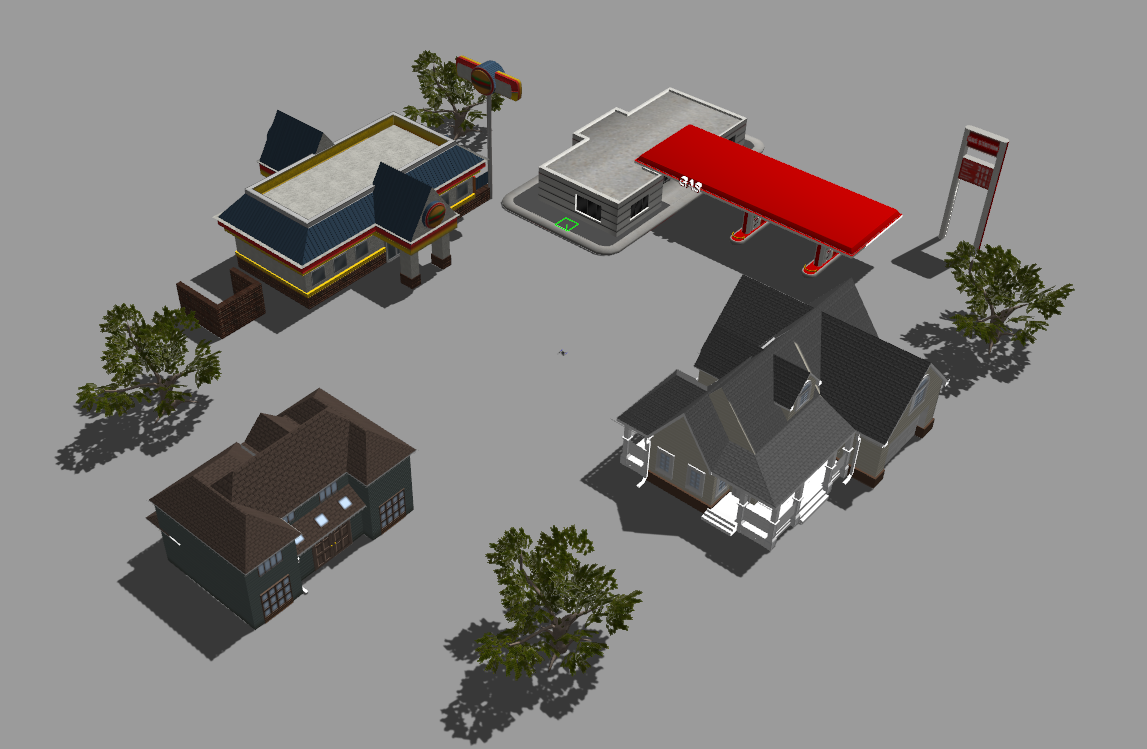}
        \caption{The training environment for 3D navigation}
        \label{fig_training_environment}
\end{figure}

In this work, an end-to-end policy network is proposed for UAV navigation in unknown 3D environment. The network is trained using a off-policy model-free DRL method. To speed up the training process, a novel framework which combines the advantages of imitation learning and reinforcement learning is proposed. Specifically, both Q-value and policy network are trained in the imitation phase and a decayed imitation loss is used to get a smooth transition between imitation and reinforcement learning phase. The training environment is shown in Fig. \ref{fig_training_environment}.

\section{Related Work}

\subsection{Learning-based UAV Navigation}
To address the UAV navigation problem with DRL method, many works only focus on the 2D situation. Ross \emph{et al} \cite{ross2013learning} proposed a vision-based navigation system for an UAV using imitation learning. However, this method needs human in the loop during the training phase. Pham \emph{et al} \cite{pham2018autonomous} train a quad-rotor to learn to navigate to the target point using a PID assisted Q-learning algorithm in an unknown environment. However, there is no obstacle in the environment. In Wang \emph{et al} 's work \cite{wang2019autonomous}, the navigation problem is formulated as a partially observable Markov decision process (POMDP) and solved by a novel online DRL method. Singla \emph{et al} \cite{singla2019memory} used the GAN architecture for depth prediction from RGB image and augmenting DRL with memory networks and temporal attention facilitates the agent to retain vital information gathered from the past observations.

Because of the difficulty, only a small amount of work focus on the 3D navigation problem. Sharma \emph{et al} \cite{sharma2012autonomous} proposed an RL based autonomous waypoint generation strategy (AWGS) for on-line path planning in unknown 2D and 3D environments. However, the policy is learned from scratch which is time consuming.

\begin{figure*}[t]
        \centering
        \includegraphics[width=16cm]{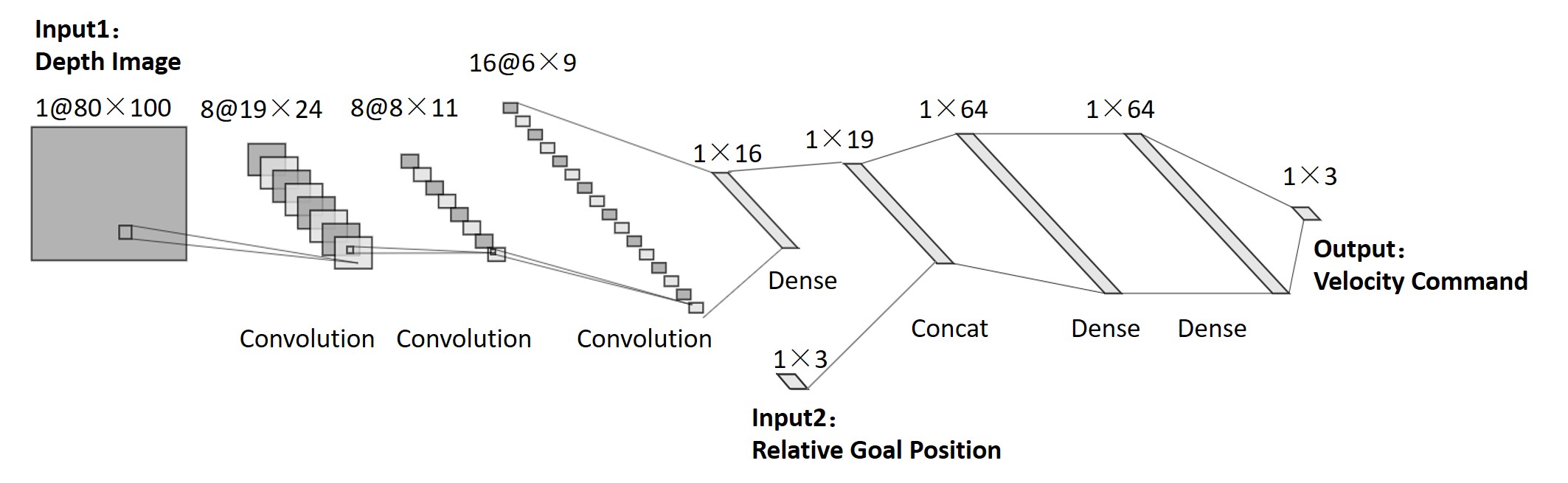}
        \caption{The policy network.}
        \label{fig:policy network}
\end{figure*}

\subsection{Learn from Demonstrations}
Demonstrations are widely used in high-dimensional robotic problems. Hester \emph{et al} proposed Deep Q-learning from Demonstrations (DQfD), that leverages small sets of demonstration data to accelerate the learning process. Vecerik \emph{et al} \cite{vecerik2017leveraging} proposed a general and model-free approach which build upon the DDPG algorithm to use demonstrations. Both demonstrations and actual interactions are used to fill the replay buffer and sampled via a prioritized replay mechanism. Similar to DQfD, DDPGfD also uses a mix of 1-step and n-step return losses and L2 regularization losses. Nair \emph{et al} \cite{nair2018overcoming} also proposed a method which builds on top of DDPG, they use BC loss and Q-Filter as an auxiliary loss function when updating the policy in the training phase. Gao \emph{et al} \cite{gao2018reinforcement} proposed Normalized Actor-Critic (NAC) which is robust to sub-optimal demonstrations. 

In application, to address the mapless navigation problem for the mobile robot, Xie \emph{et al} \cite{xie2018learning} proposed Assisted DDPG, where a classical controller is used as an alternative and switchable policy to speed up the training of DRL. This method needs the assisted controller always online in the training phase. Pfeiffer \emph{et al} \cite{pfeiffer2018reinforced} leverage prior expert demonstrations to pre-train the policy and then use a safety constrained RL method to improve the performance. However, only the policy network is pre-trained using the demonstration data, value network is still initialized randomly. When it starts interacting with the environment, the policy performance will drop because of the incorrect value function estimation. 

\section{Background}

\subsection{Reinforcement Learning for Navigation Problem}

In this work, the navigation and obstacle avoidance problem is formulated with standard Markov Decision Process (MDP) which can be solved using DRL. An MDP is defined by a tuple $<S,A,R,P,\gamma>$, which consists of a set of states $S$, a set of actions $A$, a reward function $R(s,a)$, a transition function $P(s'|s,a)$, and a discount factor $\gamma$. In each state $s \in S$, the agent takes an action $a \in A$. By executing the action $a$ in the environment, the agent receives a reward $R(s,a)$ and reaches a new state $s'$, determined from the probability distribution $P(s'|s,a)$. The goal of DRL is to find a policy $\pi$ mapping states to actions that maximizes the expected discounted total reward over the agent's lifetime. This concept is formalized by the action value function: $Q^{\pi}(s,a)=\mathbb{E}^{\pi} \left[\sum^{T}_{t=0} \gamma^{t}R(s_t,a_t)\right]$, where $\mathbb{E}^{\pi}$ is the expectation over the distribution of the admissible trajectories $(s_0,a_0,s_1,a_1,\dots)$ obtained the policy $\pi$ starting from $s_0=s$ and $a_0=a$. 

In the UAV navigation and obstacle avoidance problem, state $s$ is represented with the relative goal position and sensor data. In our case, the raw depth image obtained from a depth camera or binocular camera is used to extract the obstacle information. Action $a$ generated from the policy network $\pi(s)$ which consists of linear velocity in x, y-axis and the rotation speed in the z-axis to navigate the UAV working in 3D environment. The policy network is shown in Fig. \ref{fig:policy network}.

\subsection{Twin Delayed DDPG}
Our method builds upon an off-policy model-free reinforcement learning algorithm, Twin Delayed DDPG (TD3) \cite{fujimoto2018addressing}. A common failure mode for DDPG is that the learned Q-function begins to dramatically overestimate Q-values, which then leads to the policy breaking. TD3 addresses this issue by introducing three critical tricks: clipped double Q-Learning, delayed policy update and target policy smoothing \cite{SpinningUp2018}.

\textbf{Target policy smoothing:} Actions used to form the Q-learning target are based on the target policy, $\pi_{\theta_{\text{targ}}}$, but with clipped noise added on each dimension of the action. After adding the clipped noise, the target action is then clipped to lie in the valid action range (all valid actions $a$ satisfy $a_{Low} \leq a \leq a_{High}$). The target actions are thus:
\begin{equation}\label{eq_traget_action_noise}
        a'(s') = \text{clip}\left(\pi_{\theta_{\text{targ}}}(s') + \text{clip}(\epsilon,-c,c), a_{Low}, a_{High}\right)
\end{equation}
where $\epsilon \sim \mathcal{N}(0, \sigma)$. Target policy smoothing essentially serves as a regularize for the algorithm. It addresses a particular failure mode that can happen in DDPG: if the Q-function network develops an incorrect sharp peak for some actions, the policy will quickly exploit that peak and then have brittle or incorrect behaviour. This can be averted by smoothing out the Q-function over similar actions, which target policy smoothing is designed to do.

\textbf{Clipped double-Q learning:} TD3 concurrently learns two Q-functions, $Q_{\phi_1}$ and $Q_{\phi_2}$, by mean square Bellman error minimization, in almost the same way that DDPG learns its single Q-function. Both Q-functions use a single target, calculated using whichever of the two Q-functions gives a smaller target value:
\begin{equation}\label{eq_q_get_target}
        y(r,s',d) = r + \gamma (1-d) \min_{i=1,2} Q_{\phi_{i, \text{targ}}}(s',a'(s'))
\end{equation}
and then the parameters of both Q-value functions $\phi_1$ and $\phi_2$ are updated by one step of gradient descent using:
\begin{equation}\label{eq_update_q_network}
    \nabla_{\phi_i} \frac{1}{\mathcal{B}} \sum_{(s,a,s',r,d) \in \mathcal{B}} (Q_{\phi_i}(s,a)-y(r,s',d))^2
\end{equation}
where $i=1,2$ and $\mathcal{B}$ is a mini-batch sampled from the replay buffer $\mathcal{D}$. Using the smaller Q-value for the target, and regressing towards that, helps decrease overestimation in the Q-function.

\textbf{Delayed policy updates:} Lastly, the parameter of the policy network $\pi_{\theta}$ is updated by one step of gradient ascent to maximize the Q-value using:
\begin{equation}\label{eq_update_policy}
    \nabla_{\theta} \frac{1}{\mathcal{B}} \sum_{s \in \mathcal{B}} Q_{\phi_1}(s, \pi_{\theta}(s))
\end{equation}
which is pretty much unchanged from DDPG. However, in TD3, the policy is updated less frequently than the Q-functions are. This helps damp the volatility that normally arises in DDPG because of how a policy update changes the target.

\section{Approach}
In this section, a learning from demonstration method TD3fD (TD3 from Demonstration) is proposed to address the UAV navigation problem. Our method combines reinforcement learning and imitation learning which can get better data efficiency than learning from scratch. Notably, differing from DQfD and DDPGfD, both policy and Q-value network are initialized using imitation learning. In addition, a decaying behaviour cloning loss is used at the beginning of the training phase to stabilize the training process. 

\subsection{Problems with Behaviour Cloning}
Given a set of demonstrations that contains all the transition information $(s,a,s',r,d)$ and the corresponding environment, an agent should perform appropriate actions when it starts interacting with the environment and continues to improve \cite{gao2018reinforcement}. BC method can learn the mapping between the input observations and their corresponding expert actions, but it will lead to \textit{compounding errors}, which means an early error could potentially cascade to a sequence of mistakes, especially for the long sequence decision problem. Also, the BC method cannot deal with unseen data. Because of the demonstration set is collected using expert, a classical obstacle avoidance algorithm in our case, only correct transitions are collected without any collision. So the demonstration set is a highly biased sample of the real environment. Using the off-policy RL method directly on the demonstration set will also lead to \textit{mismatching problem}.

\subsection{TD3 with Demonstrations}
To deal with the \textit{mismatching problem} and speed up the training process, our method combines BC with RL method. The whole process has two phases: imitation and reinforcement. Our work is inspired by the previous work DQfD and DDPGfD. However, differing from DDPGfD, both policy and Q-value network are trained during imitation phase. Moreover, the imitation loss is preserved at the beginning of reinforcement phase and reduces with the training process goes on. Furthermore, we don't keep the demonstration data permanently. After certain training steps, the reinforcement phase will degenerate to the original TD3. This decayed imitation loss guarantees the stability at the beginning of reinforcement phase and can lead to a smooth transfer from demonstration set to the real environment.

For the actor-critic reinforcement learning framework, if only the policy network is pre-trained using BC method, the performance will decline dramatically when it starts interacting with the environment because of the incorrect Q-value estimation. So, in our work, both Q-value and policy network are initialized using the demonstration data in the imitation phase. The imitation loss (or BC loss) is defined with:
\begin{equation}\label{eq_bc_loss}
        L_\text{BC}(s,a) =  (a - \pi_\theta(s))^2
\end{equation}
where $a$ is the expert action.

To learn both Q-value and policy network simultaneously, the imitation loss is added to equation \eqref{eq_update_policy} as an auxiliary loss and the Q-value network is updated by maximizing $Q_{\phi_1}$ as well as  minimizing $L_\text{BC}$ simultaneously:
\begin{equation}\label{eq_policy_update_pretrain_phase}
        \nabla_{\theta} \frac{1}{\mathcal{B}} \sum_{s \in \mathcal{B}} \big[Q_{\phi_1}(s, \pi_{\theta}(s))-w L_{\text{BC}}(s,a)\big]
\end{equation}
where $w$ is the weight of imitation loss. 

\begin{algorithm}
\renewcommand{\algorithmicrequire}{ \textbf{Input:}} 
\renewcommand{\algorithmicensure}{ \textbf{Output:}} 
\caption{TD3fD}
\label{alg_TD3fD}
\begin{algorithmic}[1]
\REQUIRE  
        $\mathcal{D}^{replay}$ (initialized with demonstration data set),
        policy parameters $\theta$ and Q-function parameters $\phi_1$, $\phi_2$ (random),
        set target parameters equal to main parameters: $\theta_{\text{targ}} \leftarrow \theta$,$\phi_{\text{targ}1}\leftarrow \phi_1$, $\phi_{\text{targ}2}\leftarrow \phi_2$

\textbf{Imitation phase:}
\FOR{steps $t \in{1,2,...,N_{\text{imitation}}}$}
        \STATE{Sample a mini-batch $\mathcal{B}$ from $\mathcal{D}^{replay}$}
        \STATE{Update Q-functions parameters $\phi_1$ and $\phi_2$ with equation \eqref{eq_traget_action_noise}, \eqref{eq_q_get_target} and \eqref{eq_update_q_network}}
        \IF{update policy}
                \STATE{Update policy one step of gradient ascent with equation \eqref{eq_policy_update_pretrain_phase}}
                \STATE{Update target network}
        \ENDIF
\ENDFOR

\textbf{Reinforcement phase:}
\FOR{steps $t \in{1,2,...,N_{\text{reinforcement}}}$}
        \STATE{Observe state $s$ and select action with exploration noise
        \begin{equation*}
                a=\text{clip}(\pi_{\theta}(s)+\epsilon, a_{Low}, a_{High}), \epsilon \sim N(0,\sigma^2)
        \end{equation*}}
        \STATE{Execute $a$ in the environment and get observations ${s',r,d}$}
        \STATE{Store $(s,a,s',r,d)$ in $\mathcal{D}^{replay}$, overwriting oldest transition if over capacity}
        \STATE{Sample a mini-batch $\mathcal{B}$ from $\mathcal{D}^{replay}$}
        \STATE{Update Q-functions parameters $\phi_1$ and $\phi_2$ with equation \eqref{eq_traget_action_noise}, \eqref{eq_q_get_target} and \eqref{eq_update_q_network}}
        \IF{update policy}
                \STATE{Update policy one step of gradient ascent with equation \eqref{eq_policy_update_training_phase}}
                \STATE{Update target network}
        \ENDIF
\ENDFOR

\end{algorithmic}
\end{algorithm}

After the imitation phase, a modified TD3 algorithm is used to improve the policy network to deal with unseen scenario and correct the mismatch between the demonstration set and the real environment. To get a smooth transfer from imitation to reinforcement phase, a decay factor is added to equation \eqref{eq_policy_update_pretrain_phase}:
\begin{equation}\label{eq_policy_update_training_phase}
        \nabla_{\theta} \frac{1}{\mathcal{B}} \sum_{s \in \mathcal{B}} \big[Q_{\phi_1}(s, \pi_{\theta}(s))- \lambda w L_{\text{BC}}(s,a)\big]
\end{equation}
where $\lambda$ is the decay factor calculated by: 
\begin{equation*}
    \lambda = \max\big(0,1-\frac{t}{N}\big)
\end{equation*}
where $t$ is the current time step, $N$ is the total decay step number. At the beginning $\lambda$ is equal to 1 and will gradually decrease to 0 after $N$ steps. The TD3fD algorithm is outlined in Algorithm \ref{alg_TD3fD}.

\begin{figure*}[t]
        \centering
        \begin{subfigure}[b]{0.3\linewidth}
                \includegraphics[width=\linewidth]{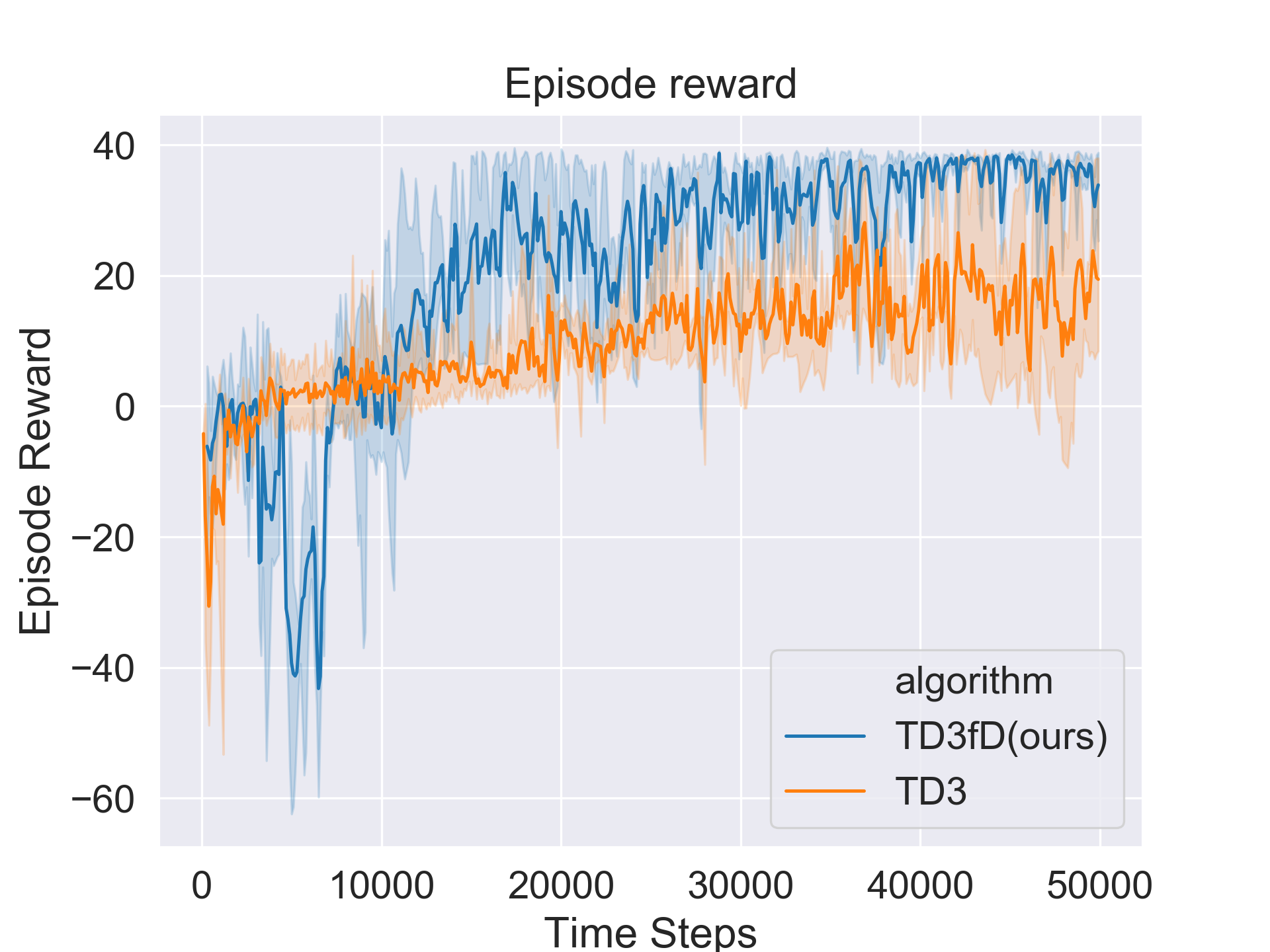}
                \caption{Episode reward}
        \end{subfigure}
        \begin{subfigure}[b]{0.3\linewidth}
                \includegraphics[width=\linewidth]{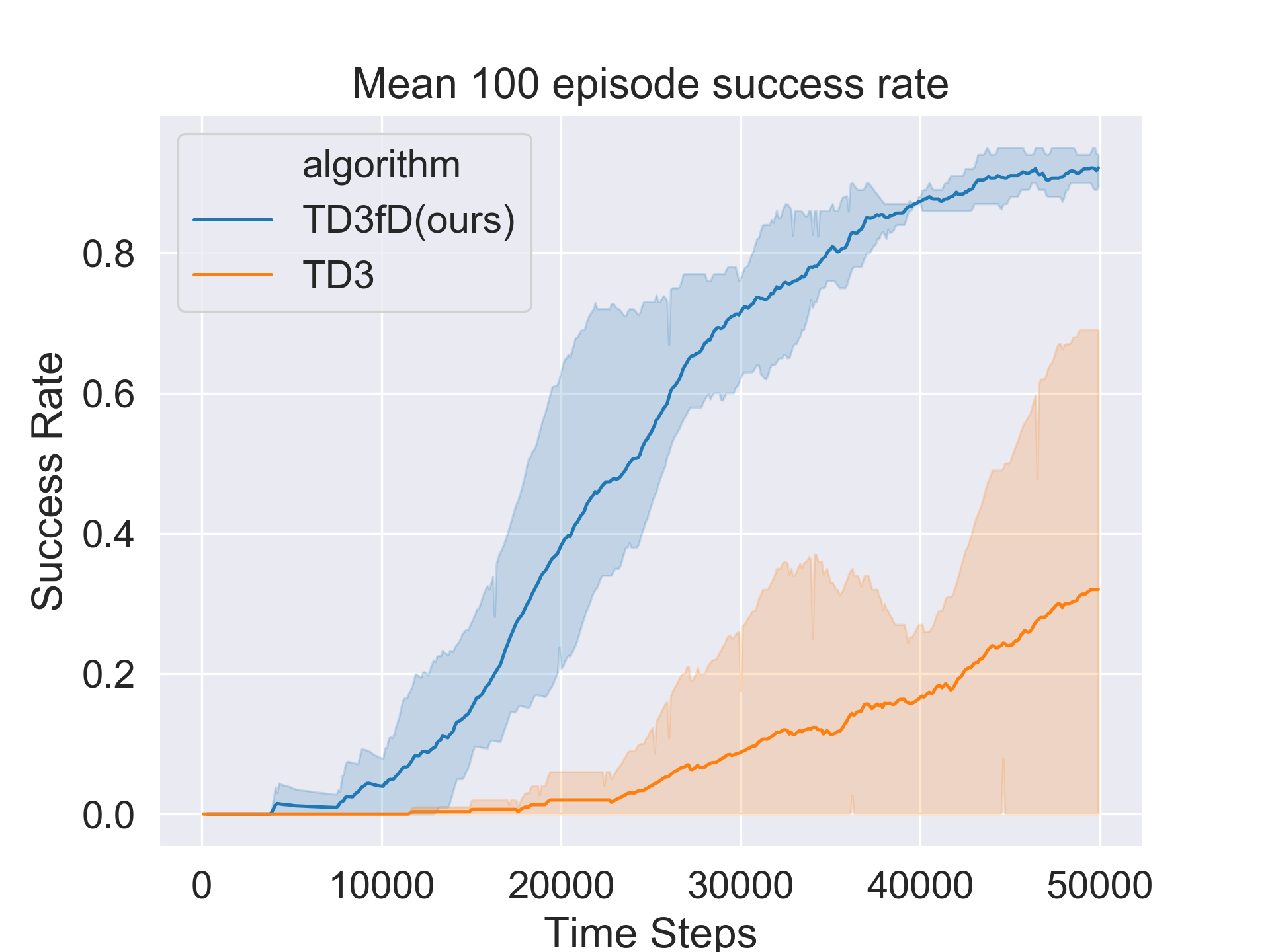}
                \caption{Mean 100 episode success rate}
        \end{subfigure}
        \begin{subfigure}[b]{0.3\linewidth}
                \includegraphics[width=\linewidth]{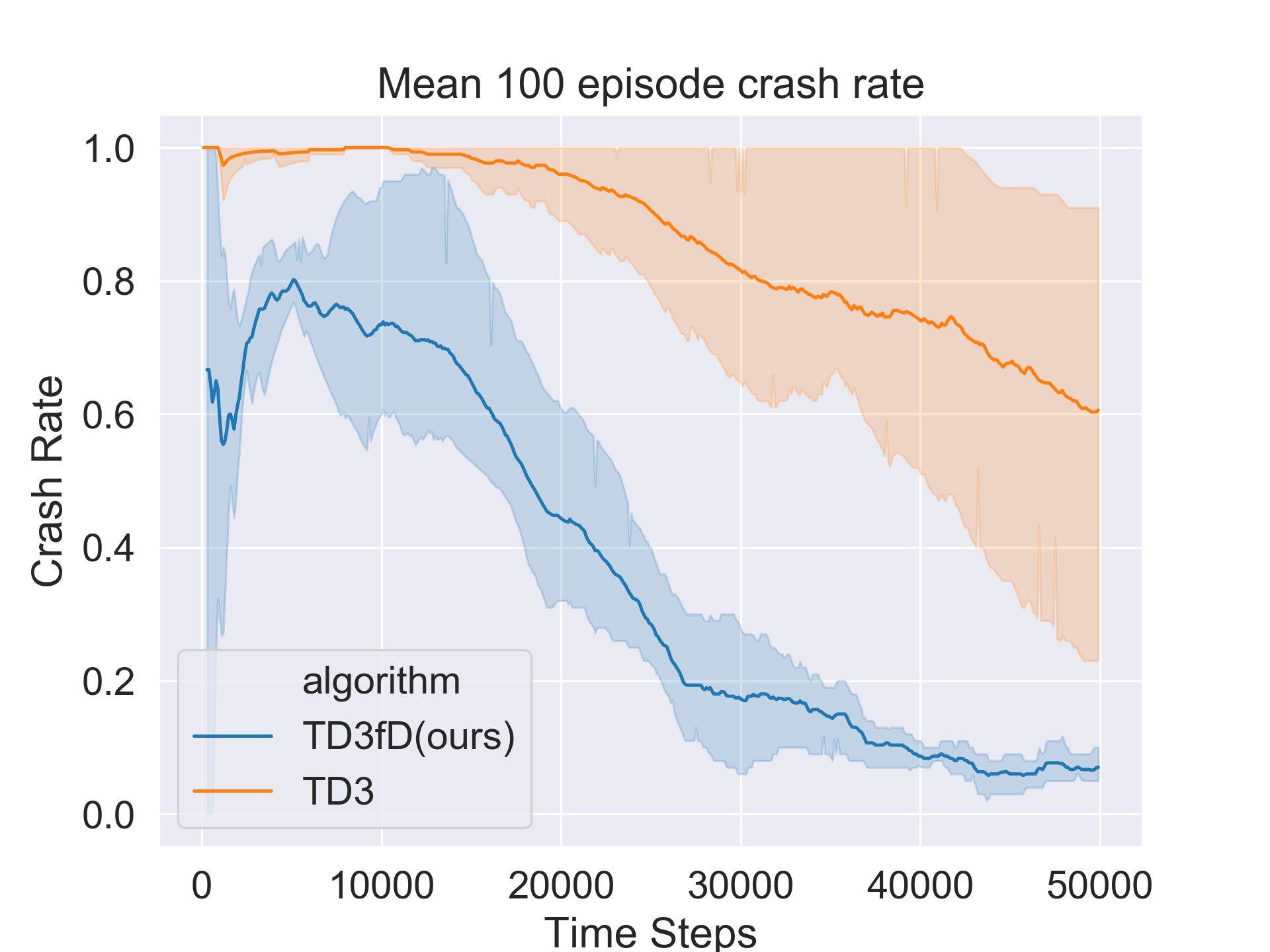}
                \caption{Mean 100 episode crash rate}
        \end{subfigure}
        \caption{Training results for 50000 time-steps over 3 random seeds.}
        \label{fig_results}
\end{figure*}

\section{Experiments}
Several experiments are conducted to evaluate the performance of the proposed TD3fD against the original TD3. The network is trained in the ROS based Gazebo simulation environment with OpenAI gym interface \cite{zamora2016extending}. In order to simulate the real-world situation as much as possible, the UAV is controlled using PX4 flight stack \cite{meier2015px4} and running in the Software In The Loop (SITL) mode. Our TD3fD algorithm is modified from the stable baseline \cite{stable-baselines} TD3 implementation which is based on OpenAI Baselines \cite{baselines}. The training environment is shown in Fig. \ref{fig_training_environment}.

\subsection{Expert Demonstration}
The PX4 local planner based on the 3DVHF* algorithm is used as the expert instead of a human. It is an open-source ROS package for obstacle avoidance. Using the depth image as input, the PX4 local planner generates a vector field histogram to represent the local information around the vehicle. Then multiple collision free trajectories are generated based on this vector field histogram and a best one is selected based on a cost function. Although this algorithm has been optimized for on-board application, it still spends much computer resource because of the look-ahead tree search algorithm.

In the expert demonstration gathering phase, 10 different goals are set randomly. The multirotor takes off at the centre of the environment and flies to the goal position guided by the local planner. The original output of the local planner is target waypoint. We transfer these target waypoints to the velocity command in UAV body frame as the expert action $a$. To get better use of the expert demonstrations, all the transition information $(s,a,s',r,d)$ is recorded and stored in the replay buffer.

\subsection{Network Framework and Training Settings}
The policy network using depth image and the relative position between the current UAV position and goal position as input. A CNN feature extractor is used to get useful information from a raw depth image. The detailed structure of the policy network is shown in Fig. \ref{fig:policy network}. The output of the policy network is velocity command, consists of forwarding speed, climbing rate and yaw rate in vehicle body frame. The activation function for the hidden layer is ReLU and tanh is used in the final dense layer to generate symmetrical control command. All commands are transformed into ROS topics and published at 5Hz. The low-level control is executed by the PX4 flight firmware. The hyperparameters of training are shown in Table \ref{table_hyperparameters}.

\begin{table}[t]
        \caption{Hyper-parameters}
        \label{table_hyperparameters}
        \begin{center}
                \begin{tabular}{cc}
                        \toprule
                        \textbf{Hyper-parameter} & \textbf{Value} \\
                        \midrule
                        mini-batch size & 128 \\ 
                        replay buffer size & 50000 \\
                        discount factor & 0.99 \\
                        learning rate & 0.0003 \\
                        soft update coefficient & 0.005 \\
                        policy update delay & 2\\
                        random exploration steps & 1000\\
                        square deviation of exploration noise & 0.1 \\
                        \bottomrule
                \end{tabular}
        \end{center}
\end{table}

\subsection{Reward Function}
The agent's objective is to reach the target in the shortest possible number of time-steps while avoiding the obstacles. The reward function provides the required feedback to the agent in the training phase. To simplify the training process, a hand-designed reward function include continuous reward is utilized:

\begin{equation}
        r(s_t)=
        \begin{cases}
                10, &\text{if success} \\
                -(d(s_t)-d(s_{t-1})) - C, &\text{otherwise}
        \end{cases}
\end{equation}
where $d(s_t)$ is the \textit{Euclidean distance} from current position to goal position at time $t$. $C$ is a constant used as time penalty. In order to reduce the variance, no punishment term is used for a crash.

\begin{figure*}[t]
        \centering
        \begin{subfigure}[b]{0.3\linewidth}
                \includegraphics[width=\linewidth]{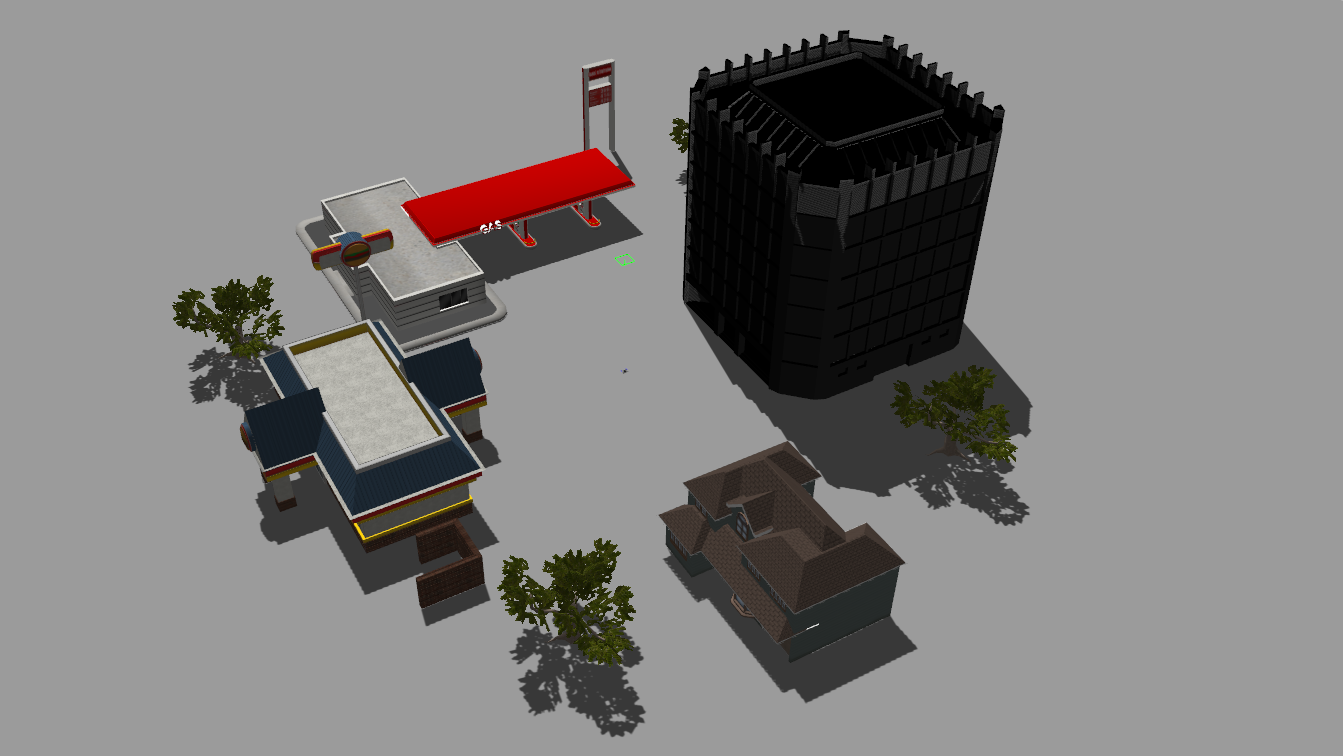}
                \caption{World2}
        \end{subfigure}
        \begin{subfigure}[b]{0.3\linewidth}
                \includegraphics[width=\linewidth]{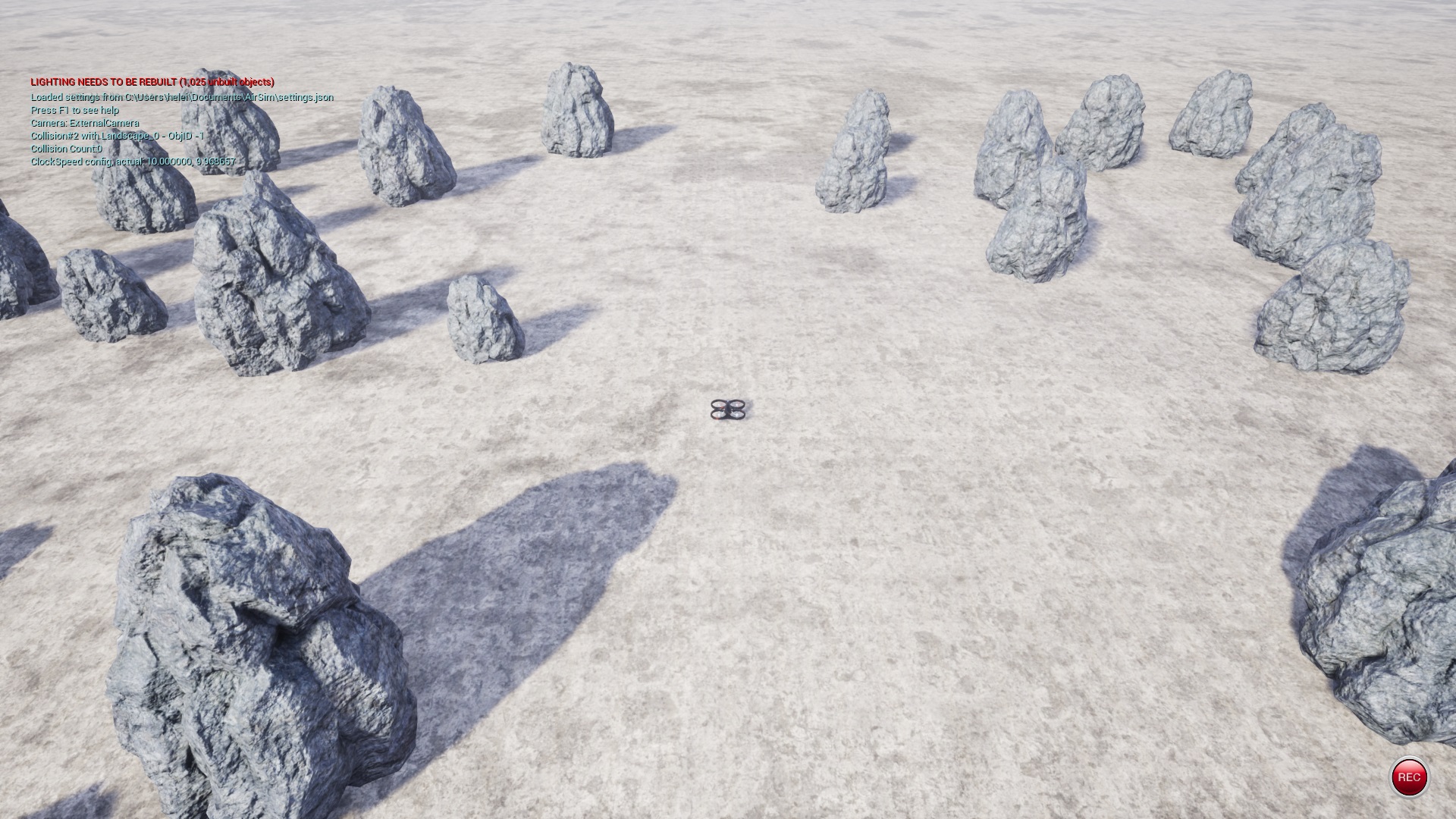}
                \caption{Rocks}
        \end{subfigure}
        \begin{subfigure}[b]{0.3\linewidth}
                \includegraphics[width=\linewidth]{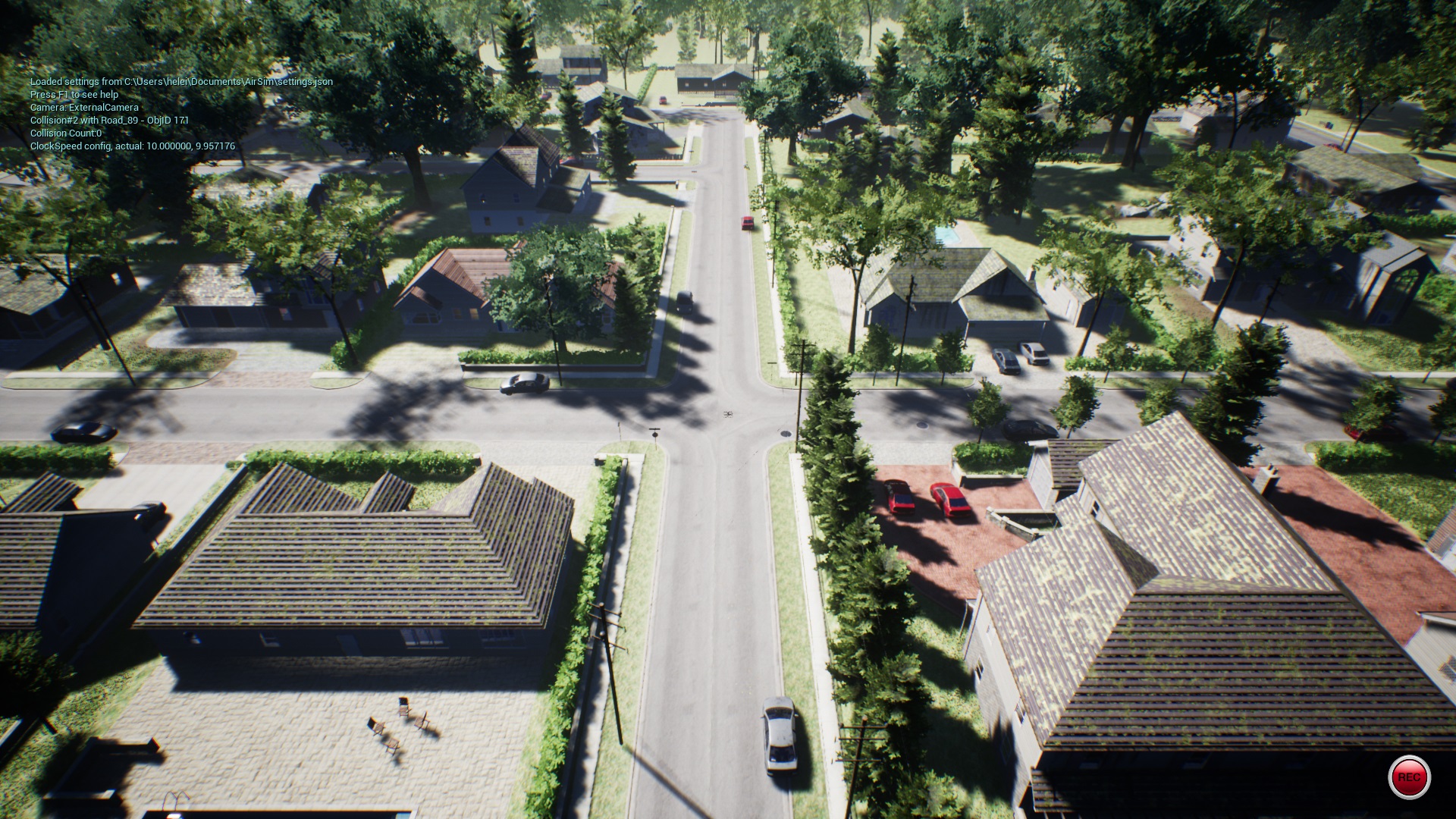}
                \caption{Neighborhood}
        \end{subfigure}
        \caption{Evaluation environments.}
        \label{fig_evaluation_environment}
\end{figure*}

\subsection{Training Results}
The imitation phase starts with the randomly initialized network and a replay buffer initialized with expert transitions. In this phase, the TD3fD algorithm is executed for 5000 time-steps with the data from replay buffer rather than interact with the environment. In our experiment, $w$ is set to 20 to get better behaviour cloning of the policy network. After imitation phase, the policy network can get some sense of the environment and can succeed occasionally. Then the training phase is executed for 50000 time-steps to improve the pre-trained network through interacting with the environment. The decay step number $N_\text{decay}$ is set to 5000.

To show the advantage of learning speed, the original TD3 algorithm is used to compare with our TD3fD method. Training results are shown in Fig. \ref{fig_results}. From the results, we can see that TD3fD learns faster than original TD3. After 50000 time-steps training, TD3fD got a acceptable success rate while the original TD3 struggled with the bad data efficiency and need more data to get the same performance. 

\subsection{Evaluation}
To test the generalization ability of the learned policy network, three new environments are built. World1 is the training environment shown in Fig. \ref{fig_training_environment}. World2 added a tall building in world1 which cannot be flown over. In addition, two environments, Rocks and Neighborhood, are built with AirSim \cite{shah2018airsim} simulator based on Unreal Engine, which can provide more visually realistic. New environments are shown in Fig. \ref{fig_evaluation_environment}.

\begin{figure}[t]
        \centering
        \includegraphics[width=6cm]{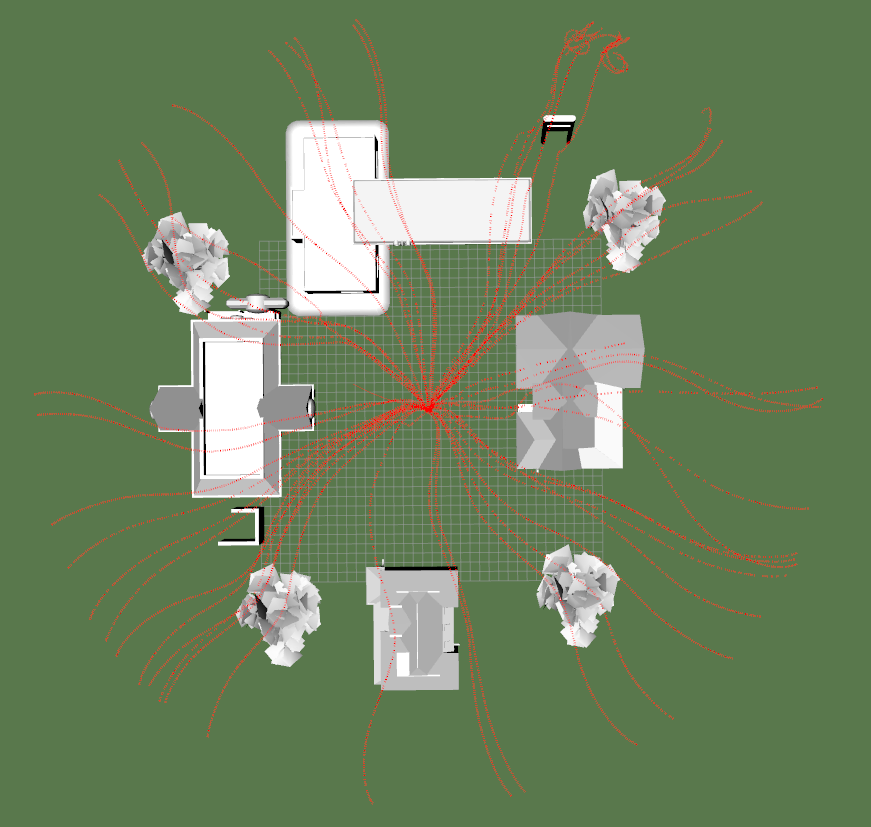}
        \caption{Trajectories generated by the learned policy using TD3fD algorithm for 50 episodes in World1.}
        \label{fig_evaluation_50}
\end{figure}

In each environment, both policy network learned using pure BC method and TD3fD are executed for 50 episodes without action noise. In World1 and World2 environment, goal position is generated randomly on a circle with a radius of 40 meters and centered on the take-off point. In the Rocks environment the radius of goal position is set to 60 meters. In the Neighborhood environment, goal position is selected randomly from a list of 10 reachable position. Trajectories generated in the training environment are shown in Fig. \ref{fig_evaluation_50}. From the trajectories we can see that the UAV learned to climb and fly over some low obstacles to reach the goal.

\begin{table}[t]
        \caption{Evaluation in different environments}
        \label{table_evaluation}
        \begin{center}
                \begin{tabular}{cccc}
                        \toprule
                        \textbf{Environment} & \textbf{Policy} & \textbf{Average reward} & \textbf{Success rate}\\
                        \midrule
                        \multirow{3}*{\makecell[c]{World1 \\ (training environment)}}
                         &expert   &32.12 & \textbf{96}$\%$  \\
                         &pure BC  &10.08 & 22$\%$  \\ 
                         &TD3fD    &\textbf{35.99} & 90$\%$  \\ 
                        \midrule
                        \multirow{3}*{\makecell[c]{World2 \\ (Gazebo)}} 
                         &expert   &\textbf{33.02} & \textbf{96}$\%$  \\
                         &pure BC  &2.79           & 16$\%$  \\ 
                         &TD3fD    &28.98          & 72$\%$  \\ 
                        \midrule
                        \multirow{2}*{\makecell[c]{Rocks \\ (AirSim)}} 
                        
                        &pure BC  &5.95 & 0$\%$  \\ 
                        &TD3fD    &56.92 & \textbf{98}$\%$  \\ 
                        \midrule
                        \multirow{2}*{\makecell[c]{Neighborhood \\ (AirSim)}} 
                        
                        &pure BC  &-10.62 & 0$\%$  \\ 
                        &TD3fD    &34.67 & \textbf{90}$\%$ \\ 
                        \bottomrule
                \end{tabular}
        \end{center}
\end{table}

The average reward and success rate for different environments are shown in Table. \ref{table_evaluation}. Because the expert controller can only run with ROS, there is no expert data in AirSim environments. From Table. \ref{table_evaluation}, the policy trained with TD3fD can greatly outperform the policy learned using pure BC method and the final performance is similar to the traditional methods. It is worth noting that, the average reward of policy learned using TD3fD exceeds the expert even with a slight low success rate, which means that the learned policy finds the shorter path than the expert. However, comparing with the training environment, the success rate declined when the learned policy is executed in the World2, because the learned policy relies too much on climb to avoid the obstacles rather than steer. It can be addressed by a better hand-designed reward function.

According to the evaluation results, the learned policy can achieve acceptable performance in different unseen environments. Because the goal distance is different with Gazebo environment, the average reward in AirSim environments cannot be compared with the gazebo environment directly. In the AirSim environment, the ground-truth state is used for low-level control. So the velocity control is better than Gazebo environment which runs all PX4 flight stack in the SITL mode including state estimation. From the success rate, we can see that the learned policy performs quite well in the complex AirSim environment, which indicates that a good state estimation is important for the obstacle avoidance and navigation problem.  

\section{Conclusions}
In this work, a DRL framework is proposed for UAV navigation and obstacle avoidance in the unknown 3D environment. Especially, expert demonstrations are used to speed up the training process and both policy and Q-value network are pre-trained in the imitation phase. Simulation results show that this learned end-to-end policy network can achieve similar performance compared with the traditional navigation method. In addition, the DRL process can be accelerated significantly leveraging only a small amount of expert demonstration. Our method shows promise for learning in the real environment and can be integrated to any other actor-critic off-policy RL method. 

While in this work, training was only sketched in the simulation environment, in future we will evaluate the learned policy in the real environment. We also plan to add some safety constraint during the training process and achieving on-policy learning in the real environment safely. 

\addtolength{\textheight}{-2cm} 

\bibliographystyle{IEEEtran}
\bibliography{./ref} 

\end{document}